\documentclass{article}
\usepackage{graphicx}
\usepackage{hyperref}
\graphicspath{ {.} }
\title{RealFace - Pedestrian Face Dataset}
\author{Leonardo Ramos Thomas}
\date{August 2024}
\usepackage[style=apa,backend=biber]{biblatex}
\begin{document}
\maketitle
\section{Introduction to The Real Face Dataset}
\paragraph{The Real Face Dataset is a pedestrian face detection benchmark dataset in the wild, comprising over 11,000 images and over 55,000 detected faces in various ambient conditions. The dataset aims to provide a comprehensive and diverse collection of real-world face images for the evaluation and development of face detection and recognition algorithms. The Real Face Dataset is a valuable resource for researchers and developers working on face detection and recognition algorithms. With over 11,000 images and 55,000 detected faces, the dataset offers a comprehensive and diverse collection of real-world face images. This diversity is crucial for evaluating the performance of algorithms under various ambient conditions, such as lighting, scale, pose, and occlusion. The dataset's focus on real-world scenarios makes it particularly relevant for practical applications, where faces may be captured in challenging environments.}
\paragraph{In addition to its size, the dataset's inclusion of images with a high degree of variability in scale, pose, and occlusion, as well as its focus on practical application scenarios, sets it apart as a valuable resource for benchmarking and testing face detection and recognition methods. The challenges presented by the dataset align with the difficulties faced in real-world surveillance applications, where the ability to detect faces and extract discriminative features is paramount.}
\paragraph{The Real Face Dataset provides an opportunity to assess the performance of face detection and recognition methods on a large scale. Its relevance to real-world scenarios makes it an important resource for researchers and developers aiming to create robust and effective algorithms for practical applications. Public access to dataset can be found here:}
\href{https://github.com/leo7r/RealFaceDataset}{RealFace Dataset}
\section{Face Detection: An Overview}
\paragraph{Face detection is a fundamental task in computer vision, with applications ranging from facial recognition to video surveillance. Accurate face detection is essential for subsequent tasks such as facial recognition, emotion analysis, and gender identification. Various datasets have been developed to aid in the development and evaluation of face detection algorithms  \parencite{pavel_korshunov_0f7a8005}.}
\paragraph{The most notable datasets in this domain include the FERET dataset and Labeled Faces in the Wild dataset . Early face datasets, like the PIE and FERET datasets, were primarily collected under controlled environments, allowing for high-performance results on constrained datasets  \parencite{tianyue_zheng_21463ba0}. However, the complexity of real-world faces poses challenges for these algorithms when applied to practical applications . To overcome these limitations, the face recognition community turned to unconstrained datasets that better reflect real-world conditions  \parencite{jennifer_newman_4ca5587b}.}
\paragraph{These unconstrained datasets, such as the Real Face Dataset, provide a more realistic representation of the challenges faced in real-world scenarios. They contain images captured in various ambient conditions, with factors such as lighting variations, occlusions, and pose variations.}
\section{Understanding the Real Face Dataset Structure}
\paragraph{The Real Face Dataset is structured using XML annotations that provide information about each detected face. These annotations specify the bounding box coordinates of each face, including the xmin, ymin, xmax, and ymax values. }
\begin{figure}
    \centering
    \includegraphics[width=1\linewidth]{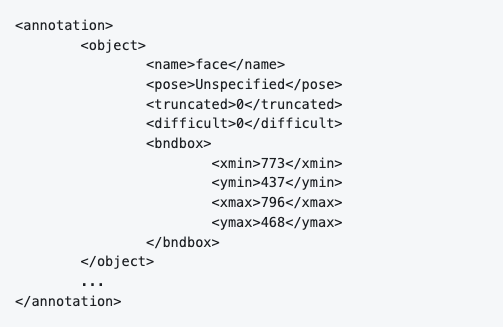}
    \caption{XML Annotation Example}
    \label{fig:enter-label}
\end{figure}
\paragraph{When it comes to annotating datasets, there are various formats used to represent the location and attributes of objects within the images. One commonly used annotation format is the Pascal VOC format \parencite{pascal_voc___cvat_3255dd78}. This format is widely adopted in the computer vision and machine learning communities due to its flexibility and support for rich annotations. The Pascal VOC format allows for the representation of object instances, keypoint annotations, and captions, making it suitable for a wide range of tasks including object detection, segmentation, and keypoint estimation.}
\paragraph{The choice of using the Pascal VOC format for the Real Face Dataset is driven by its versatility and compatibility with popular deep learning frameworks and libraries. The format's ability to capture complex object relationships and attributes aligns with the diverse and real-world nature of the face images in the dataset. Additionally, the format's standardization enables seamless integration with existing annotation tools and evaluation metrics, facilitating collaboration and comparison across different research efforts.}
\paragraph{By utilizing the Pascal VOC format for annotating the Real Face Dataset, we aim to streamline the development and evaluation of face detection and recognition algorithms. The format's comprehensive representation of object annotations empowers researchers and developers to explore the dataset's rich collection of real-world face images, ultimately advancing the state-of-the-art in face-related computer vision tasks.}
\section{Analyzing Over 11000 Images in the Real Face Dataset}
\paragraph{To analyze over 11,000 images in the Real Face Dataset, we employed a rigorous process of filtering and annotation. Firstly, we focused on analyzing public pedestrian videos, extracting frames that contained one or more faces. This initial step resulted in a large collection of frames which we then extensively revised to retain only the high-quality ones. Our thorough revision involved filtering out duplicates and eliminating frames with poor image quality or visibility of faces.}
\paragraph{After the initial filtering process, we manually annotated all the more than 55,000 faces with meticulous precision. This labor-intensive task involved marking the bounding box coordinates of each detected face, ensuring an accurate representation of their position within the images. Our meticulous annotation process aimed to capture the diversity and complexity of real-world faces, including variations in scale, pose, and occlusion.}
\paragraph{To ensure the quality and diversity of the Real Face Dataset, a key consideration is to select very diverse pedestrian videos that capture a wide range of ambient conditions. It is essential to include videos with different illuminations, diverse contexts, and varying face densities to comprehensively represent real-world scenarios. By incorporating such diversity, the dataset will encompass the challenges faced in practical applications and enable the evaluation of face detection and recognition algorithms under various conditions.}
\paragraph{Diverse pedestrian videos not only contribute to the richness of the dataset but also facilitate the enhancement of algorithms to perform robustly in real-world settings. Additionally, the inclusion of different illuminations, contextual variations, and varying face densities will allow for a comprehensive evaluation of algorithm performance across a wide spectrum of scenarios. This approach aligns with the objective of providing a valuable resource for researchers and developers working on face detection and recognition algorithms.}
\section{Diversity of Detected Faces: More than 55000 Examples}
\paragraph{The Real Face Dataset has been carefully curated to ensure the accuracy and diversity of detected faces.}
\paragraph{This dataset consists of over 55,000 manually annotated faces, capturing a wide range of ethnicities including Asian (13.43), Black or African American (5.22), Hispanic or Latino (5.97), and White (75.37).}
\paragraph{The Real Face Dataset also includes faces of various ages, genders, and facial expressions.}
\paragraph{This ensures that the dataset is representative of the real-world population and provides a robust benchmark for evaluating face detection algorithms in terms of their ability to accurately identify and classify faces from diverse backgrounds, allowing for more inclusive and equitable face recognition technologies. In addition to the diversity in detected faces, the Real Face Dataset also incorporates variations in scale, pose, and occlusion. This ensures that the dataset presents a wide range of challenges and realistic scenarios for face detection algorithms to overcome.}
\section{Ambient Conditions in the Real Face Dataset}
\paragraph{The Real Face Dataset includes images and videos captured in both indoor and outdoor environments, providing a comprehensive representation of ambient conditions.}
\begin{figure}
    \centering
    \includegraphics[width=1\linewidth]{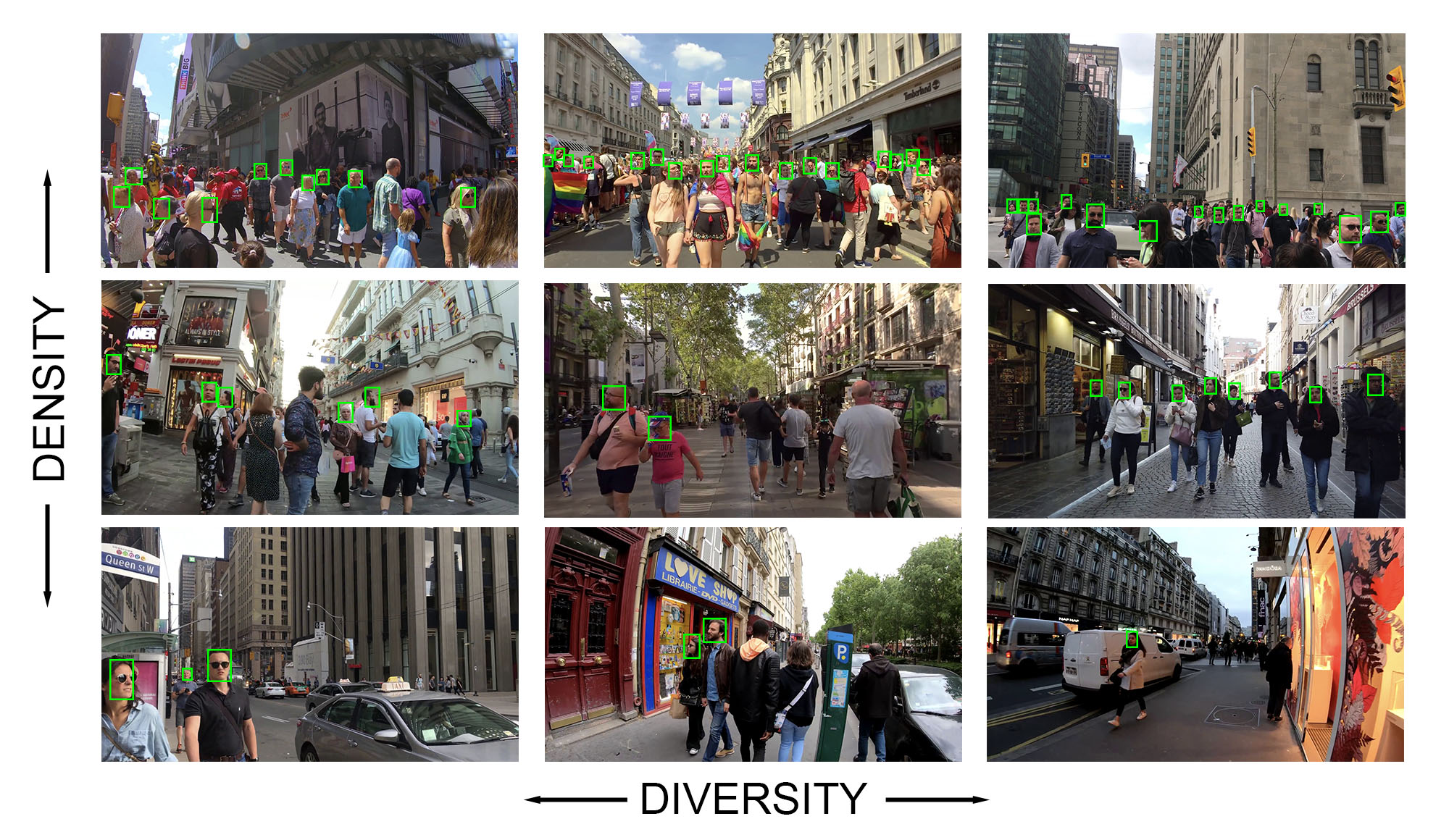}
    \caption{Density and Diversity in RealFace Dataset}
\end{figure}
\paragraph{The dataset covers a variety of lighting conditions, including natural daylight, artificial indoor lighting, and low-light or nighttime illumination. This variety of ambient conditions allows for the evaluation of face detection algorithms' robustness and adaptability in different lighting scenarios. Furthermore, the Real Face Dataset also takes into account variations in weather conditions, such as rain, snow, fog, and glare. The inclusion of these ambient conditions adds an additional layer of complexity to the dataset, challenging face detection algorithms to accurately identify and localize faces under adverse weather or environmental conditions.}
\section{Applications and Implications of The Real Face Dataset}
\paragraph{The Real Face Dataset has significant implications for various applications in the field of face detection and recognition. One application is in the development and evaluation of face-detection algorithms  \parencite{yunqiu_wang_42668695} for surveillance systems. These algorithms can be used to automatically detect and track faces in large-scale gathering environments, such as airports, stadiums, or crowded public spaces. Another application is in the development of facial recognition systems  \parencite{xavier_fontaine_faf06323} for identity verification and access control. These systems rely on accurate face detection as a crucial first step in the recognition process. By providing a diverse and challenging dataset, the Real Face Dataset enables researchers and developers to design and evaluate robust face detection algorithms that can effectively handle variations in pose, lighting, and environmental conditions commonly encountered in real-world scenarios. Additionally, the Real Face Dataset has implications for the development of face anti-spoofing techniques. With the increasing prevalence of deepfake technology  \parencite{zuheng_ming_b20e2bdf} and face spoofing attacks, it is crucial to develop robust methods for distinguishing between real faces and fake faces.}
\paragraph{The Real Face Dataset can serve as a valuable resource for training and evaluating face anti-spoofing algorithms, as it includes instances of occlusion, head pose variations, different scales, face masks, and various lighting conditions. Furthermore, the Real Face Dataset can also be utilized for research purposes, such as studying facial expression recognition  \parencite{alaa_s__al_waisy_fd265423} in real-world scenarios. The inclusion of diverse facial expressions in the dataset allows for the development and evaluation of algorithms that can accurately recognize and interpret facial expressions in different contexts.}
\paragraph{These applications and implications highlight the importance of datasets like the Real Face Dataset in advancing the field of face detection and recognition.}
\section{Challenges and Limitations of The Real Face Dataset}
\paragraph{While the Real Face Dataset provides a valuable resource for face detection and recognition research, it does have some challenges and limitations. One challenge is the potential bias in the dataset due to the representation of certain demographics. To mitigate this, efforts should be made to ensure a diverse and representative sample of individuals across various age groups, ethnicities, and genders in the dataset. Another limitation of the Real Face Dataset is that it primarily focuses on pedestrian faces in outdoor environments. This limits its applicability to other domains, such as indoor surveillance or specific occupational environments. Furthermore, the Real Face Dataset may not fully capture all possible variations in real-world conditions, as environmental factors can be highly diverse and unpredictable. Additionally, the Real Face Dataset may not include certain uncommon scenarios or extreme conditions that could potentially occur in real-world situations. }
\paragraph{The Real Face Dataset serves as an important benchmark dataset in the field of face anti-spoofing and face detection. However, it is crucial for future research to continue expanding and diversifying datasets in order to tackle the challenges posed by emerging technologies and evolving real-world scenarios}
\section{Future Directions in Pedestrian Face Detection Research}
\paragraph{Future directions in pedestrian face detection research should focus on addressing the limitations of existing datasets, including the Real Face Dataset.}
\paragraph{Efforts should be made to create datasets that are more inclusive and representative of diverse demographics, as well as datasets that encompass a wider range of environments and scenarios. Furthermore, the incorporation of more challenging factors into the datasets, such as occlusions, varying lighting conditions, and complex backgrounds, would help to improve the performance of face detection algorithms in real-world scenarios. }
\paragraph{Moreover, research should also explore the integration of multi-modal data sources, such as utilizing video-based datasets or incorporating 3D facial information, to enhance the robustness and accuracy of pedestrian face detection algorithms. By addressing these limitations and expanding the scope of datasets, researchers can develop more reliable and effective face-detection algorithms that are capable of handling the complexities of real-world environments and scenarios.}
\section{Conclusion: The Real Face Dataset and Beyond}
\paragraph{Overall, the Real Face Dataset has contributed significantly to the field of face detection in real-world scenarios. However, it is important to acknowledge its limitations and the need for further advancements in dataset creation. Researchers should continue to push the boundaries of dataset diversity, incorporating more challenging factors and expanding into multi-modal data sources. These advancements will enable the development of more robust and accurate pedestrian face detection algorithms, ultimately improving the performance of face recognition systems in practical applications. }
\paragraph{Furthermore, collaboration among researchers and the sharing of datasets and algorithms are crucial for the progress of pedestrian face detection research. In today's rapidly changing world, the significance of accurate pedestrian face detection algorithms cannot be overstated. In order to meet the demands of practical applications and real-world scenarios, future research should focus on improving dataset diversity, incorporating complex factors such as occlusions, varying lighting conditions, and complex backgrounds. Furthermore, the integration of multi-modal data sources and the exploration of cross-modality training approaches would further enhance the robustness and accuracy of pedestrian face detection algorithms. }
\paragraph{In conclusion, the Real Face Dataset has been a valuable resource for studying pedestrian face detection in real-world scenarios. It has provided researchers with a large-scale dataset containing diverse and challenging images, allowing for the development and evaluation of face detection algorithms in realistic conditions. These advancements will ultimately lead to more reliable and effective face detection algorithms that can be applied in a wide range of real-world applications, such as access control systems, surveillance, and human-computer interaction.}
\printbibliography
\end{document}